\documentclass[conference,letterpaper,fleqn]{IEEEtran}
\usepackage[cmex10]{amsmath}
\usepackage{amssymb}
\usepackage{mathtools}
\usepackage{bm}
\usepackage[pdftex]{graphicx}
\usepackage[font=footnotesize]{subfig}
\usepackage[svgnames]{xcolor}
\usepackage{cite}
\usepackage{flushend}
\usepackage{hyperref}
\hypersetup{%
  pdftitle={Adiabatic Quantum Computing for Binary Clustering},
  pdfauthor={Christian Bauckhage},
  pdfsubject={quantum computing, machine learning},
  pdfkeywords={adiabatic quantum computing, binary clustering, Ising model},
  colorlinks=true,
  urlcolor=blue,
  citecolor=green,
  bookmarks=false}

\newcommand{\PY}{\emph{Python }}

\newcommand{\QT}{\emph{QuTiP }}

\newcommand{\mat}[1]{\boldsymbol{{#1}}}

\renewcommand{\vec}[1]{\boldsymbol{{#1}}}

\newcommand{\ipt}[2]{#1^T #2}
\newcommand{\nrm}[1]{\bigl \lVert #1 \bigr \rVert^2}

\newcommand{\dsq}[2]{\bigl \lVert #1 - #2 \bigr \rVert^2}

\newcommand{\amin}[1]{\operatorname*{argmin}_{#1}}
\newcommand{\amax}[1]{\operatorname*{argmax}_{#1}}

\newcommand{\ket}[1]{\vert {#1} \rangle}
\newcommand{\Ket}[1]{\big\vert {#1} \big\rangle}

\graphicspath{{./}}

\begin{document}

\bstctlcite{IEEEexample:BSTcontrol}

\title{Adiabatic Quantum Computing \\ for Binary Clustering}

\author{%
\IEEEauthorblockN{Christian Bauckhage, Eduardo Brito, Kostadin Cvejoski, Cesar Ojeda, Rafet Sifa, Stefan Wrobel}
\IEEEauthorblockA{Fraunhofer IAIS, Sankt Augustin, Germany \\ B-IT, University of Bonn, Bonn, Germany}}

\maketitle

\begin{abstract}
Quantum computing for machine learning attracts increasing attention and recent technological developments suggest that especially adiabatic quantum computing may soon be of practical interest. In this paper, we therefore consider this paradigm and discuss how to adopt it to the problem of binary clustering. Numerical simulations demonstrate the feasibility of our approach and illustrate how systems of qubits adiabatically evolve towards a solution.
\end{abstract}

\section{Introduction}

Quantum computing promises fast solutions to a wide range of optimization problems and thus holds considerable potential for machine learning \cite{Schuld2014-AIT,Wittek2014-QML,Biamonte2016-QML}. However, while the quantum machine learning literature so far mainly focused on the quantum gate paradigm, noticeable technological progress leading to commercial devices is happening in adiabatic quantum computing \cite{Bian2010-TIM,Johnson2011-QAW}. 

Current adiabatic quantum computers are geared towards solving quadratic unconstrained binary optimization problems or Ising models. A simple strategy for setting up established learning algorithms to run on such devices is therefore to attempt to (re-)formulate or approximate their minimization or maximization objectives in terms of Ising models. In this paper, we apply this strategy to a simple unsupervised learning problem, namely binary clustering.

Since we attempt to present our ideas in a manner that is accessible to a wider audience, we discuss basic concepts of machine learning and quantum computing alike. In particular, we structure our presentation into two major parts: 
\begin{enumerate}
\item In sections~\ref{sec:k-means} and \ref{sec:2-means}, we discuss an alternative objective function for $k$-means clustering and demonstrate that it allows for expressing the problem of $k=2$-means clustering in terms of an Ising model that should allow for implementation on a D-Wave computer \cite{Bian2010-TIM,Johnson2011-QAW}
\item In sections~\ref{sec:AQC} and \ref{sec:examples}, we show how to set our model up for adiabatic quantum computation and present practical examples which illustrate its use for clustering. Using the \PY toolbox \QT \cite{Johansson2013-QUT}, we simulate adiabatic evolutions of qubit registers on a digital computer. This allows for visualizing the behavior of the amplitudes of basis states and thus provides insights into the inner workings of the approach proposed in this paper. 
\end{enumerate}
 
First, however, we specify our problem setting and introduce the notation we will use throughout.

\section{Problem Setting and Notation}

The practical problem we address in this paper is that of binary clustering of numerical data. Accordingly, we let
\begin{equation}
X = \bigl\{ \vec{x}_1, \vec{x}_2, \ldots, \vec{x}_n \bigr\}
\end{equation}
denote a finite sample of real valued data vectors. These are to be clustered into two nonempty subsets $X_1$ and $X_2$ such that $X_1 \cap X_2 = \emptyset$ and $X_1 \cup X_2 = X$. The two resulting cluster means or cluster centroids therefore amount to
\begin{equation}
\vec{\mu}_i = \frac{1}{n_i} \sum_{\vec{x} \in X_i} \vec{x}
\end{equation}
where $n_i = \lvert X_i \rvert$. 

We further assume the data in $X$ to be of zero mean. This does not cause loss of generality and is a common prerequisite in machine learning practice. For our particular scenario we note the following implication
\begin{equation}
\label{eq:zeromean}
\tfrac{1}{n} (n_1 \, \vec{\mu}_1 + n_2 \, \vec{\mu}_2) = \vec{0} \; \Leftrightarrow \; n_1 \, \vec{\mu}_1  = - n_2 \, \vec{\mu}_2.
\end{equation} 
Thus, the centroids of any two clusters $X_1$ and $X_2$ contained in a set of zero mean data must either both coincide with the zero vector or will necessarily be of opposite sign.

\section{An Alternative $k$-Means Objective Function} 
\label{sec:k-means}

$K$-means clustering is a popular prototype-based clustering technique used to partition a data set $X$ into $k$ disjoint clusters $X_1, \ldots, X_k$ each of which is defined as 
\begin{equation}
X_i = \left\{ \vec{x} \in X \; \middle| \; \dsq{\vec{x}}{\vec{\mu}_i} < \dsq{\vec{x}}{\vec{\mu}_j} \; \forall \, i \not= j \right\}.
\end{equation}
Given this definition, the problem at the heart of $k$-means clustering is to find appropriate cluster prototypes. Well known algorithms such as those of Lloyd \cite{Lloyd1982-LSQ}, Hartigan \cite{Hartigan1979-AAS}, MacQueen \cite{MacQueen1967-MFC}, or derivations thereof accomplish this by minimizing the overall within cluster scatter
\begin{align}
\label{eq:objective1}
S_W(k) & = \sum_{i = 1}^k \, \sum_{\vec{x} \in X_i} \, \dsq{\vec{x}}{\vec{\mu}_i}
\end{align}
with respect to the cluster centroids $\vec{\mu}_1, \ldots, \vec{\mu}_k$.

Since the minimization objective in \eqref{eq:objective1} explicitly involves the distances $\lVert \vec{x} - \vec{\mu}_i \rVert$ that occur in the definition of a cluster, it formalizes an intuitive idea. Yet, despite its seeming simplicity, $k$-means clustering proves to be NP hard \cite{Aloise2009-NPH}. Algorithms such as the ones in \cite{Lloyd1982-LSQ,Hartigan1979-AAS,MacQueen1967-MFC} are therefore but heuristics for which there is no guarantee that they will find the global minimum of \eqref{eq:objective1}. 

This stirred interest in quantum computing implementations in particular of Lloyd's algorithm which can yield logarithmic speed-up \cite{Aimeur2007-QCA,Aimeur2013-QSU,Lloyd2013-QAF,Wiebe2015-QAF}. Consequently, multiple runs, i.e.~searches for appropriate minima, can be expected to be carried out efficiently once corresponding computers become available. 

Adiabatic quantum computing implementations based on Ising models, on the other hand, seem not to have been reported yet. This is likely because $k$-means clustering is often taken to be synonymous with Lloyd's algorithm for which an Ising model reformulation is difficult to conceive.  Next, we therefore recall an alternative objective function for $k$-means clustering and demonstrate how it leads to an Ising model.

A curiously little used fact is that the problem of finding cluster centroids by minimizing \eqref{eq:objective1} is equivalent to the problem of finding cluster centroids by maximizing the following weighted sum of pairwise distances or overall between cluster scatter
\begin{equation}
\label{eq:objective2}
S_B(k) = \sum_{i,j=1}^k n_i n_j \dsq{\vec{\mu}_i}{\vec{\mu}_j}.
\end{equation}

As we show in the appendix, this can easily be seen using Fisher's analysis of variance \cite{Fisher1921-OTP}. Next, however, we demonstrate how \eqref{eq:objective2} leads to an Ising model for $k=2$ means clustering.

\section{An Ising Model for $k=2\,$ Means Clustering} 
\label{sec:2-means}

\begin{figure}[t!]
\centering
\subfloat[$n=500$ data points sampled from two bivariate Gaussians]{\includegraphics[width=0.8\columnwidth]{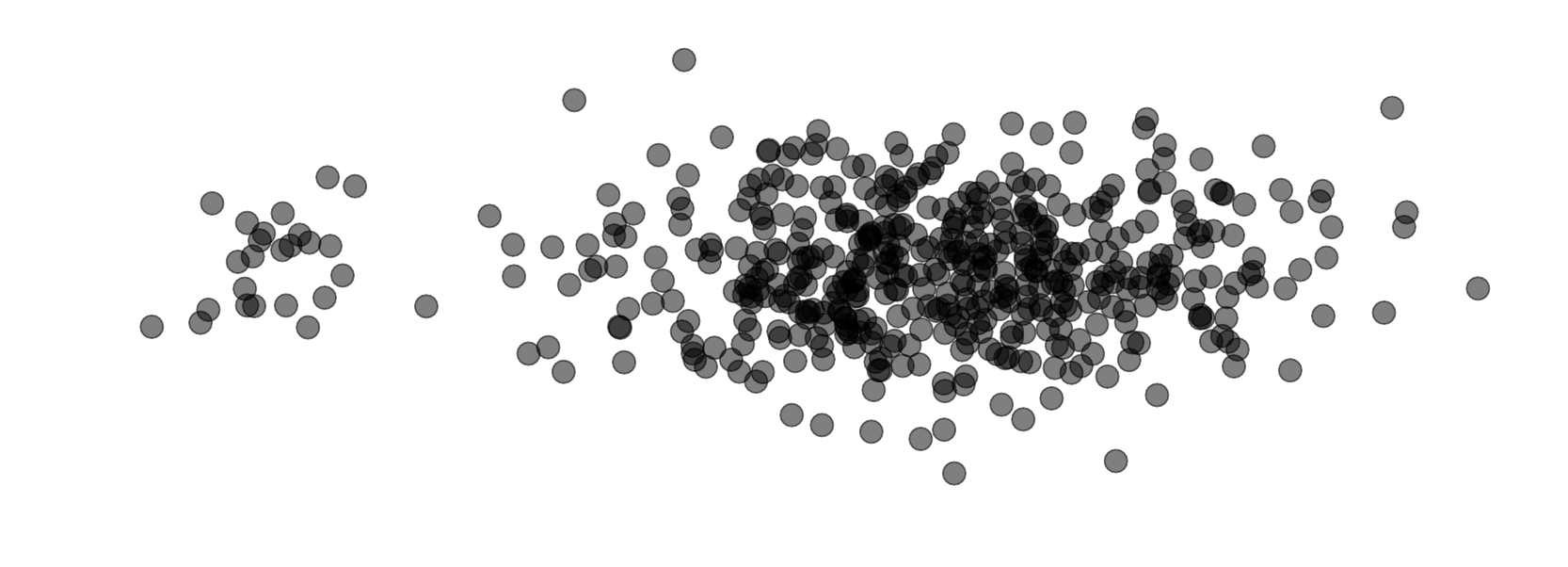}} 

\subfloat[$k=2$ clusters produced by Lloyd's algorithm]{\includegraphics[width=0.8\columnwidth]{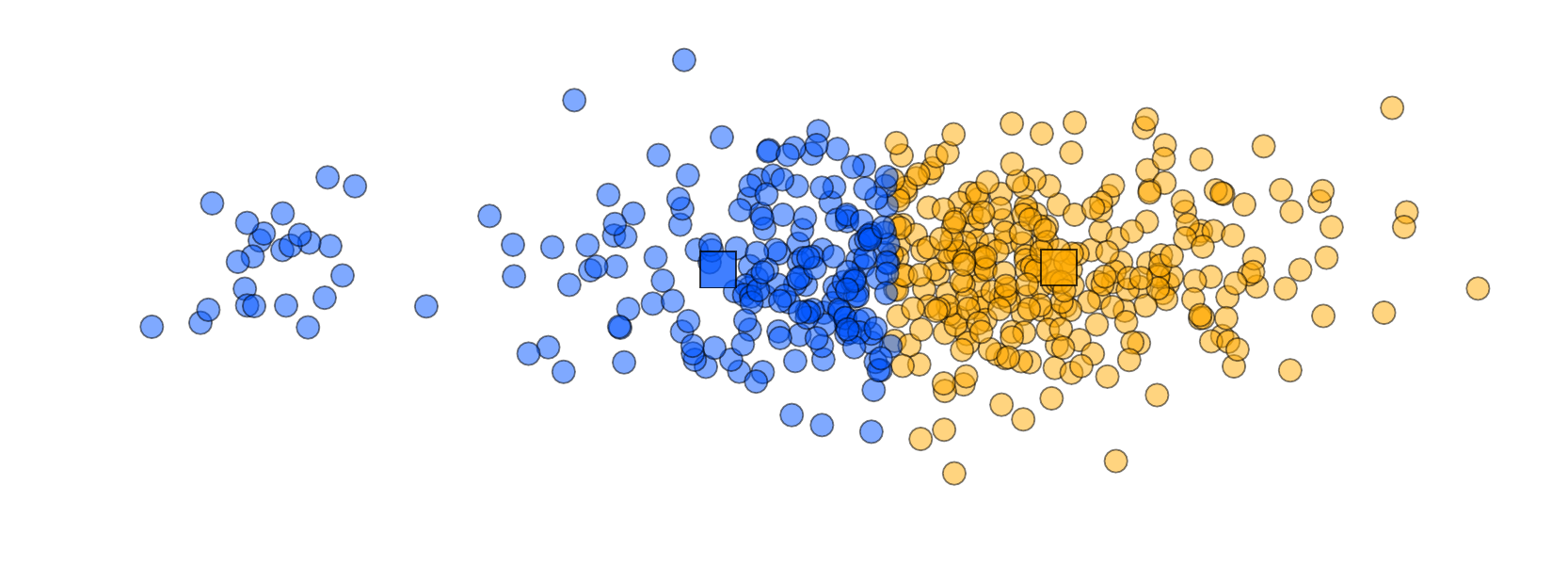}}
\caption{\label{fig:kmDidactic} Didactic illustration of the fact that Lloyd's procedure for $k$-means clustering (a.k.a~\emph{the} $k$-means algorithm) may produce clusters of comparable sizes even is the data actually consists of clusters of different sizes. It is agnostic of cluster shapes and distances \cite{MacKay2003-ITI}.}
\end{figure}

For the special case of $k=2$, the maximization objective in \eqref{eq:objective2} simplifies to
\begin{equation}
\label{eq:objective3}
S_B(2) = 2 \, n_1 n_2 \, \dsq{\vec{\mu}_1}{\vec{\mu}_2}
\end{equation} 

Interestingly, this expression provides an intuition as to why $k$-means clustering often tends to produce clusters of about equal size even if the given data contain clusters of unequal sizes (see the didactic example in Fig.~\ref{fig:kmDidactic}). 

In order for $S_B(2)$ to be large, both the distance $\lVert \vec{\mu}_1 - \vec{\mu}_2 \rVert$ between the cluster centers and the product $n_1 n_2$ of the cluster sizes have to be large. However, since the sum $n_1 + n_2 = n$ of sizes is fixed, their product will be maximal if $n_1 = n_2 = \tfrac{n}{2}$.

This observation provides a heuristic handle to rewrite the maximization objective in \eqref{eq:objective3}. Assuming that at a solution we will likely have $n_1 \approx n_2 \approx \frac{n}{2}$ allows for the approximation
\begin{align}
2 \, n_1 n_2 \, \dsq{\vec{\mu}_1}{\vec{\mu}_2} 
& \approx 2 \, \tfrac{n^2}{4} \, \dsq{\vec{\mu}_1}{\vec{\mu}_2} \\
& = 2 \, \bigl \lVert \tfrac{n}{2} \bigl(\vec{\mu}_1 - \vec{\mu}_2 \bigr) \bigr \rVert^2 \\
& = 2 \, \dsq{n_1 \, \vec{\mu}_1}{n_2 \, \vec{\mu}_2} \label{eq:heuristic}.
\end{align}

Using this heuristic, the problem of $k=2$ means clustering then becomes to maximizes the squared Euclidean norm in \eqref{eq:heuristic}. This, however, constitutes a quadratic unconstrained binary optimization problem that is equivalent to an Ising model. In order to see why, we next express the norm in \eqref{eq:heuristic} in a form that does not explicitly depend on the $\vec{\mu}_i$. 

To this end, we collect the given data in an $m \times n$ data matrix $\mat{X} = [\vec{x}_1, \vec{x}_2, \ldots, \vec{x}_n]$ and introduce two binary indicator vectors $\vec{z}_1, \vec{z}_2 \in \{0,1\}^n$ which indicate cluster memberships in the sense that entry $l$ of $\vec{z}_i$ is $1$ if $\vec{x}_l \in X_i$ and $0$ otherwise. This way, we can write 
\begin{align}
n_1 \, \vec{\mu}_1 & = \mat{X} \vec{z}_1 \\
n_2 \, \vec{\mu}_2 & = \mat{X} \vec{z}_2
\end{align}
and therefore
\begin{equation}
\label{eq:bipolar}
\dsq{n_1 \, \vec{\mu}_1}{n_2 \, \vec{\mu}_2} = \nrm{\mat{X} \bigl( \vec{z}_1 - \vec{z}_2 \bigr)} = \nrm{\mat{X} \vec{s}}.
\end{equation}

Note that the vector $\vec{s}$ we introduced in \eqref{eq:bipolar} is guaranteed to be a bipolar vector $\vec{s} \in \{-1,1\}^n$ because, in hard $k$-means clustering, every given data point is assigned to one and only one cluster so that 
\begin{equation}
\vec{z}_1 - \vec{z}_2 = \vec{z}_1 - (\vec{1} - \vec{z}_1) = 2 \, \vec{z}_1 - \vec{1} \in \{-1,1\}^n.
\end{equation}
But this is to say that we have found a simple Ising model for binary clustering of zero mean data. On the one hand, since 
\begin{equation}
\lVert \mat{X} \vec{s} \rVert^2 = \vec{s}^T \mat{X}^T \mat{X} \vec{s} = \vec{s}^T \mat{Q} \, \vec{s},
\end{equation}
the problem of maximizing the norm in \eqref{eq:heuristic} is equivalent to solving
\begin{equation}
\label{eq:Ising1}
\amin{\vec{s} \in \{-1,1\}^n} \, - \sum_{i,j=1}^n Q_{ij} \, s_i \, s_j.
\end{equation}
Because of \eqref{eq:zeromean}, on the other hand, the solution will necessarily be a vector $\vec{s}$ whose entries are not all equal and therefore induce a bipartition of the data in $\mat{X}$. 

Looking at \eqref{eq:Ising1}, three remarks appear to be in order. First of all, matrix $\mat{Q}$ is a Gram matrix where $Q_{ij} = \ipt{\vec{x}_i}{\vec{x}_j}$. Since the given data therefore enters the problem only in form of inner products, the clustering criterion derived in this section allows for invoking the kernel trick \cite{Schoelkopf2002-LWK} and is thus applicable to wide variety of practical problems.

Second of all, \eqref{eq:Ising1} exposes the ``hardness'' of $k=2$ means clustering for it reveals it as an integer programming problem. That is, it shows that binary clustering is to find an appropriate label vector $\vec{s} \in \{-1,1\}^n$ whose entries assign data points to clusters. A na\"ive solution would therefore be to evaluate \eqref{eq:Ising1} for each of the $2^n$ possible assignments of $n$ data points to $2$ clusters. On a classical computer this is clearly impractical for $n \gg 1$. On an adiabatic quantum computer, however, we can prepare a system of $n$ qubits in a superposition of $2^n$ states each of which reflects a possible solution. Given appropriate Hamiltionians, the system can then be evolved such that, when measured, it will likely collapse to a state that corresponds to a good solution.

Third of all, we observe a form of symmetry because, if $\vec{s}$ solves \eqref{eq:Ising1}, then so does $-\vec{s}$ since we did not specify whether an entry of, say, $+1$ is supposed to indicate membership to cluster one or two. To remove this ambiguity, we may remove a degree of freedom from our model. W.l.o.g.~we can, for instance, fix $s_n = +1$ and solve \eqref{eq:Ising1} for the remaining $n-1$ entries of $\vec{s}$. This way, the problem becomes to solve
\begin{equation}
\label{eq:Ising2}
\amin{\vec{s} \in \{-1,1\}^{n-1}} \, - \sum_{i,j=1}^{n-1} Q_{ij} \, s_i \, s_j - 2 \sum_{j=1}^{n-1} Q_{nj} \, s_j - Q_{nn}
\end{equation}
which we recognize as yet another Ising energy minimization problem.

\section{Adiabatic Quantum Binary Clustering}
\label{sec:AQC}

Next, we summarize basic ideas behind adiabatic quantum computing and how to set the above Ising models for binary clustering correspondingly. Quantum computing experts may safely skip this section.

When using the model in \eqref{eq:Ising1} to perform adiabatic quantum clustering of $n$ data points into $2$ clusters, we consider a system of $n$ qubits that is in a superposition of $2^n$ basis states
\begin{equation}
\label{eq:system}
\Ket{\psi(t)} = \sum_{i=0}^{2^n -1} a_i(t) \, \Ket{\psi_i} 
\end{equation}
where the time dependent coefficients or amplitudes $a_i \in \mathbb{C}$ obey $\sum_i \lvert a_i \rvert^2 = 1$. We understand each of the different basis states 
\begin{align}
\Ket{\psi_0} & = \Ket{000 \ldots 000} \\
\Ket{\psi_1} & = \Ket{000 \ldots 001} \\
\Ket{\psi_2} & = \Ket{000 \ldots 010} \\
\Ket{\psi_3} & = \Ket{000 \ldots 011} \\
& \;\; \vdots \notag
\end{align}
as an indicator vector that represents one of the $2^n$ possible assignment of $n$ points to $2$ clusters and use the common shorthand to express tensor products, for instance
\begin{equation}
\Ket{\psi_1} = \Ket{000 \ldots 001} = \Ket{0} \otimes \Ket{0} \otimes \ldots \otimes \Ket{1}.
\end{equation}

If the system in \eqref{eq:system} evolves under the influence of a time-dependent Hamiltonian $H(t)$, its behavior is governed by the Schr\"odinger equation
\begin{equation}
\frac{\partial}{\partial t} \, \Ket{\psi(t)} = -i \, H(t) \, \Ket{\psi(t)}
\end{equation}
where we have set $\hbar = 1$. Adiabatic quantum computing is concerned with evolutions like this and makes use of the adiabatic theorem \cite{Born1928-BDA}. It states that if a quantum system starts out in the ground state of a Hamiltonian operator which then gradually changes, the system will end up in the ground state of the resulting Hamiltonian. To harness this for problem solving, one prepares a system to begin in the ground state of a problem independent Hamiltonian $H_B$ and adiabatically evolves it towards a Hamiltonian $H_P$ whose ground state represents a solution to the problem at hand \cite{Farhi2000-QCB,Lucas2014-IFO,Albash2016-AQC}. 

For the problem of binary clustering, we therefore consider periods ranging from $t=0$ to $t=\tau$ and let the Hamiltonian at time $t$ be a convex combination of two static Hamiltonians, namely
\begin{equation}
H(t) = \left( 1 - \tfrac{t}{\tau} \right) H_B + \tfrac{t}{\tau} H_P.
\end{equation}
  
In order to set up a specific problem Hamiltonian for \eqref{eq:Ising1}, we follow standard suggestions \cite{Farhi2000-QCB,Lucas2014-IFO,Albash2016-AQC} and simply define 
\begin{equation}
H_P = - \sum_{i,j = 1}^n Q_{ij} \, \sigma_z^i \, \sigma_z^j 
\end{equation}
where $\sigma_z^i$ denotes the Pauli spin matrix $\sigma_z$ acting on the $i$th qubit, that is
\begin{equation}
\sigma_z^i = \underbrace{I \otimes I \otimes \ldots \otimes I}_{i-1 \text{\,terms}} \otimes \, \sigma_z \otimes \underbrace{I \otimes I \ldots \otimes I}_{n-i \text{\, terms}}.
\end{equation}

Again following standard suggestions we then choose $H_B$ to be orthogonal to $H_P$, for instance
\begin{equation}
H_B = - \sum_{i=1}^n \sigma_x^i
\end{equation}
where $\sigma_x^i$ is defined as above, this time with respect to the Pauli spin matrix $\sigma_x$.

When working with the Ising model in \eqref{eq:Ising2}, we proceed similarly. Here, we consider a system of $n-1$ qubits and set up the problem and beginning Hamiltonian as
\begin{align}
H_P & = - \sum_{i,j = 1}^{n-1} Q_{ij} \, \sigma_z^i \, \sigma_z^j - 2 \sum_{j = 1}^{n-1} Q_{nj} \, \sigma_z^j \, I^{\otimes (n-1)}
\intertext{and}
H_B & = - \sum_{i=1}^{n-1} \sigma_x^i
\end{align}
respectively.

To compute a clustering, we then let the corresponding qubit system $\ket{\psi(t)}$ evolve from $\ket{\psi(0)}$ to $\Ket{\psi(\tau)}$ where $\ket{\psi(0)}$ is chosen to be the ground state of $H_B$. That is, if $\lambda$ denotes the smallest eigenvalue of $H_B$, the initial state $\ket{\psi(0)}$ of the system corresponds to the solution of
\begin{equation}
H_B \Ket{\psi(0)} = \lambda \, \Ket{\psi(0)}.
\end{equation}

\begin{figure*}[t!]
\begin{minipage}[b]{0.28\textwidth}
\subfloat[$n=8$ data points in $\mathbb{R}^2$]{\includegraphics[width=\textwidth]{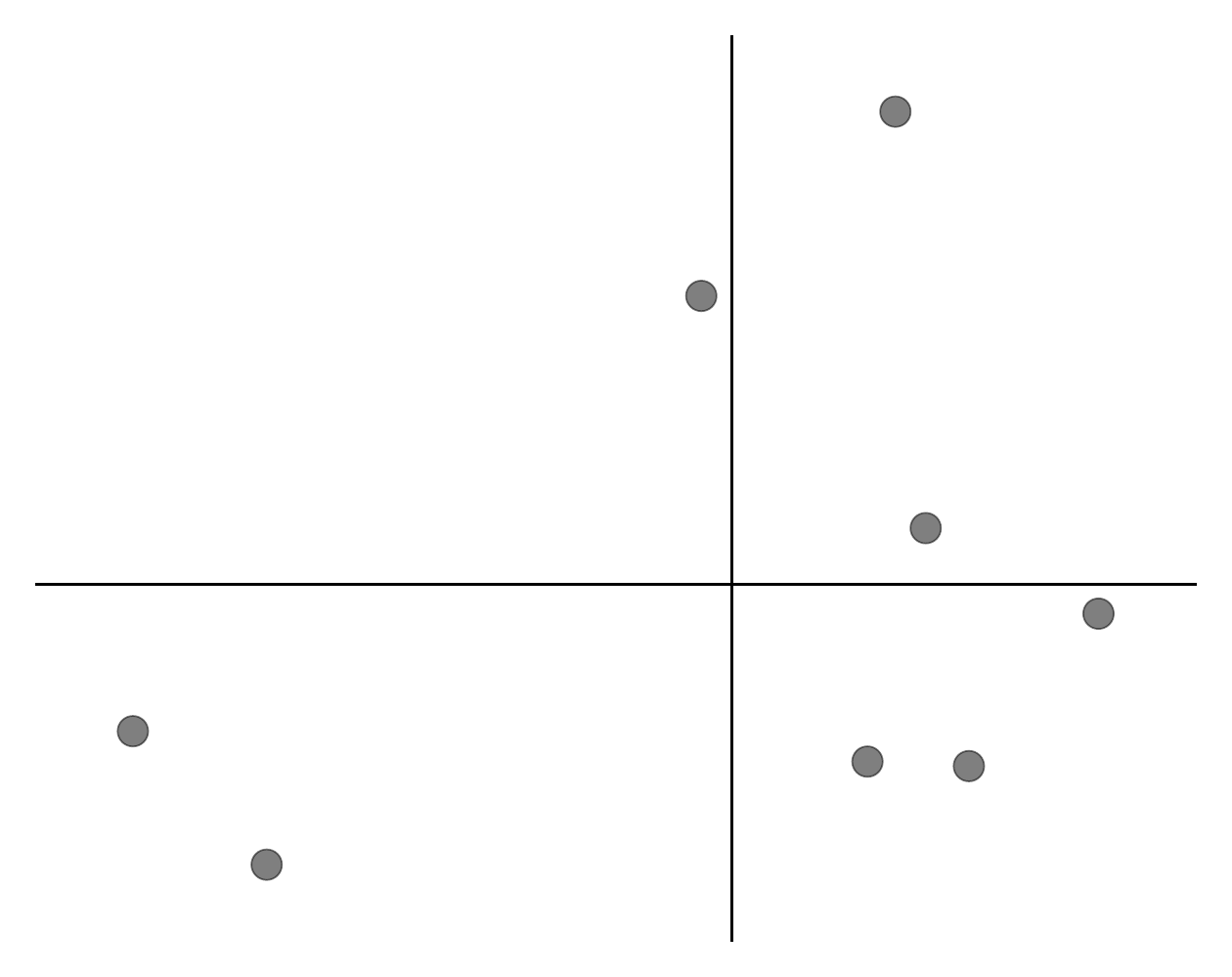}} \\
\subfloat[binary clustering result]{\includegraphics[width=\textwidth]{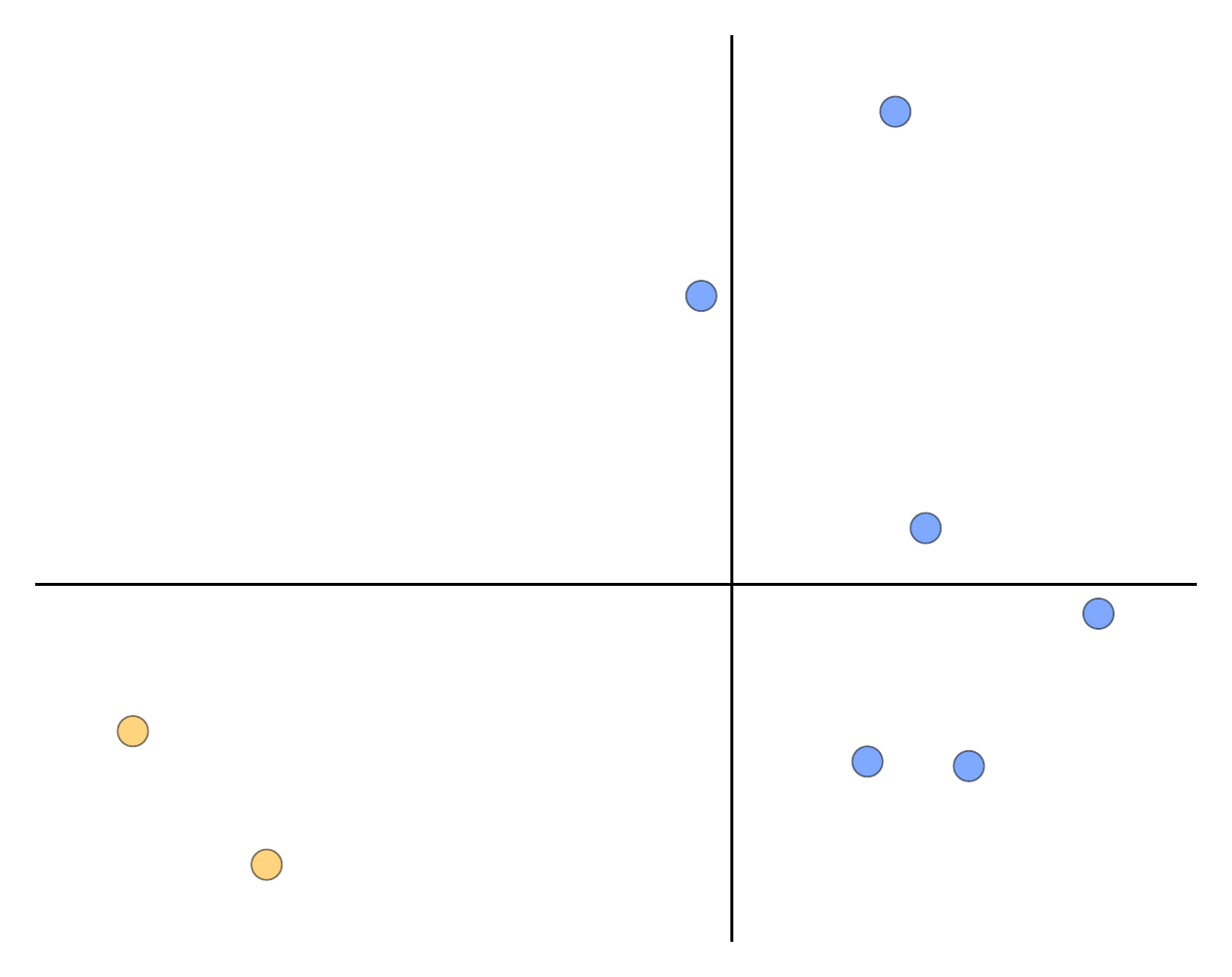}}
\end{minipage}
\hfill
\begin{minipage}[b]{0.65\textwidth}
\subfloat[adiabatic evolution of the amplitudes of the basis states $\Ket{\psi_i}$]{\includegraphics[width=\textwidth]{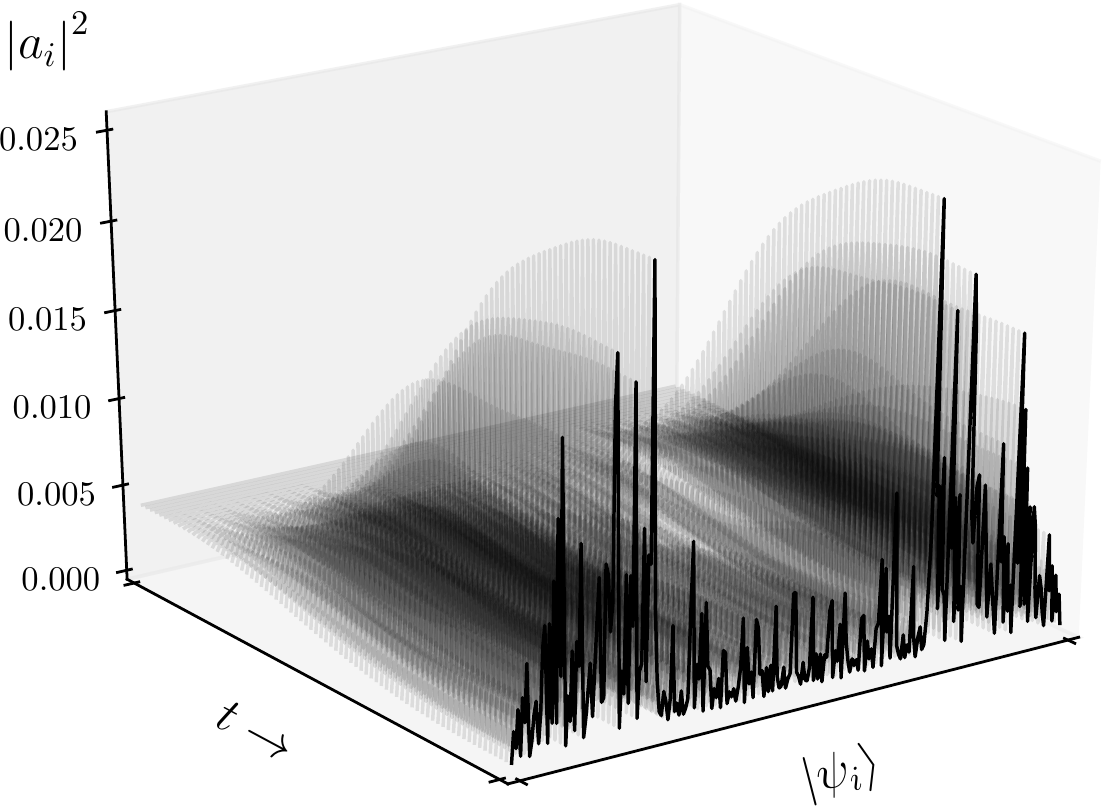}}
\end{minipage}
\caption{\label{fig:example1} Didactic example of adiabatic quantum binary clustering using the Ising model in \eqref{eq:Ising1}. (a) a zero mean sample of $8$ data points; (b) the corresponding clustering result; (c) adiabatic evolution of a corresponding system of $8$ qubits. During its evolution over time $t$, the system is in a superposition of $2^8 = 256$ basis states $\Ket{\psi_i}$ each of which represents a possible binary clustering of the data. At the beginning of the process, it is equally likely to find the system in any of these states. At the end of the process, two basis states have noticeably higher amplitudes $\lvert a_i \rvert^2$ than the others and are therefore more likely to be measured; these are $\Ket{00111111}$ and $\Ket{11000000}$ and they both induce the clustering shown in (b).}
\end{figure*}

Finally, upon termination of the adiabatic evolution, a measurement is performed on the $n$ qubit system. This will cause the wave function $\Ket{\psi(\tau)}$ to collapse to a particular basis state and the probability for this state to be $\Ket{\psi_i}$ is given by the amplitude $\lvert a_i(\tau) \rvert^2$. However, since the adiabatic evolution was steered towards the problem Hamiltonian $H_P$, basis states that correspond to ground states of $H_P$ are more likely to be found.

On an adiabatic quantum computer, all these components of the proposed binary clustering algorithm can be prepared correspondingly and the adiabatic evolution be carried out physically. On a classical digital computer, we may simulate this process by numerically solving
\begin{equation}
\label{eq:numeric}
\Ket{\psi(\tau)} = -i \int_0^\tau H(t) \, \Ket{\psi(t)} \, dt.
\end{equation}
This is the approach we will consider in the next section.

\begin{figure*}[t!]
\begin{minipage}[b]{0.28\textwidth}
\subfloat[$n=8$ data points in $\mathbb{R}^2$ one of which has been preassigned to a cluster]{\includegraphics[width=\textwidth]{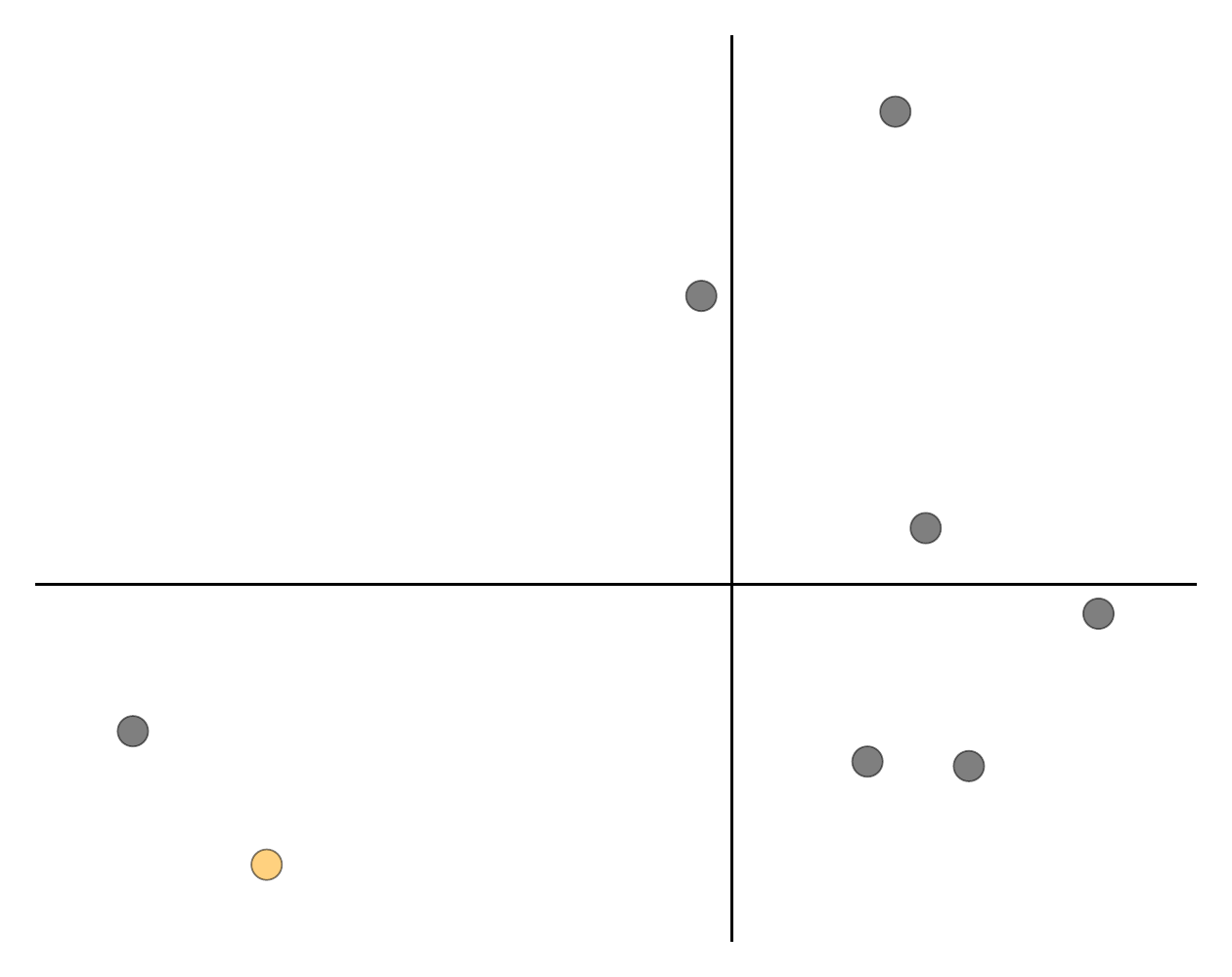}} \\
\subfloat[binary clustering result]{\includegraphics[width=\textwidth]{blobs2-2-6-aqc-result.pdf}}
\end{minipage}
\hfill
\begin{minipage}[b]{0.65\textwidth}
\subfloat[adiabatic evolution of the amplitudes of $2^7$ states $\Ket{\psi_i}$]{\includegraphics[width=\textwidth]{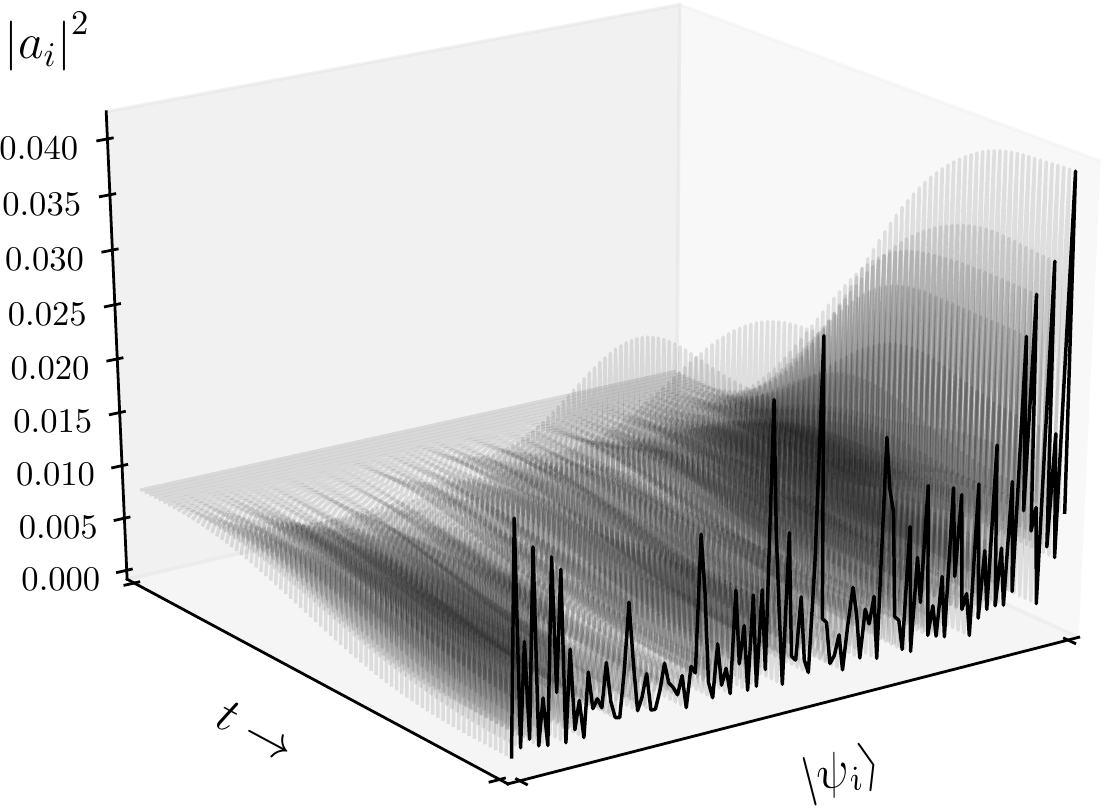}}
\end{minipage}
\caption{\label{fig:example2} Didactic example of adiabatic quantum binary clustering using the Ising model in \eqref{eq:Ising2}. (a) a zero mean sample of $8$ data points one which has been manually preassigned to a cluster; (b) the corresponding clustering result; (c) adiabatic evolution of a system of $7$ qubits. During its evolution over time $t$, the system is in a superposition of $2^7 = 128$ basis states $\Ket{\psi_i}$ each of which represents a possible binary clustering of the $7$ unlabeled data points. At the beginning of the process, it is equally likely to find the system in any of these states. At the end of the process, one of them has a higher amplitude $\lvert a_i \rvert^2$ than the others and is therefore more likely to be measured; this state is $\Ket{0000001}$ and it induces the clustering in (b).}
\end{figure*}

\section{Practical Examples}
\label{sec:examples}

In this section, we present two practical examples which demonstrate the feasibility of the proposed approach. Both examples are rather didactic and mainly intended to illuminate the process of the quantum adiabatic evolution of systems of $n$ or $n-1$ qubits, respectively. 

In both our examples, we consider the same data matrix $\mat{X} = [\mat{X}_1, \mat{X}_2] \in \mathbb{R}^{2 \times n}$ whose column vectors form two clusters of size $n_1$ and $n_2$, respectively and evolve an $n$ or $n-1$ qubit system in order to uncover these clusters. We deliberately restrict ourselves to a rather small $n = n_1 + n_2$ as this allows us to comprehensible visualize the evolution of the corresponding $2^n$ or $2^{n-1}$ basis states. We simulate both quantum adiabatic evolutions on a digital computer and use the \PY quantum computing toolbox \QT \cite{Johansson2013-QUT} in order to numerically solve \eqref{eq:numeric}.

Figure~\ref{fig:example1}(a) shows a zero mean sample of $n=8$ data points. In order to quantum adiabatically cluster these data according to the Ising model in \eqref{eq:Ising1}, we consider a system of $8$ qubits. The problem Hamiltonian $H_P$, the beginning Hamiltonian $H_B$, and the initial state $\Ket{\psi(0)}$ of this system are prepared as discussed in section~\ref{sec:AQC}. Given these components, we then solve \eqref{eq:numeric} for $t \in [0, \tau=75]$.

Figure~\ref{fig:example1}(c) illustrates the temporal evolution of the amplitudes $\lvert a_i(t) \rvert^2$ of the $2^8 = 256$ basis states $\Ket{\psi_i}$ the quantum system $\Ket{\psi(t)}$ can be in. At $t=0$, all states are equally likely but over time their amplitudes begin to increase or decrease. At $t=\tau$, two of the basis states have considerably higher amplitudes than the others so that a measurement will likely cause the system to collapse to either of these more probable states. These two basis states are $\Ket{00111111}$ and $\Ket{11000000}$ and can be understood as cluster indicator vectors which both produce the result in Fig.~\ref{fig:example1}(b).

The visualization of the quantum adiabatic evolution in Fig.~\ref{fig:example1} confirms our remarks in section~\ref{sec:2-means} where we noted that the Ising model in \eqref{eq:Ising1} is ambiguous in that it has two equally valid solutions. Our second experiment therefore investigates the use of the Ising model in \eqref{eq:Ising2} which we proposed as a way of avoiding this ambiguity.

Figure~\ref{fig:example2}(a) shows the same zero mean sample of $n=8$ data points as in our first example. This time, however, one of them has already been assigned to a cluster. This way, a degree of freedom of the clustering problem has been removed and we may consider the Ising model in \eqref{eq:Ising2} to cluster the data. We therefore prepare a system of $n-1$ qubits, the problem Hamiltonian $H_P$, the beginning Hamiltonian $H_B$, and the initial state $\Ket{\psi(0)}$ as discussed above and again solve \eqref{eq:numeric} for $t \in [0, \tau=75]$.

Figure~\ref{fig:example2}(c) illustrates the temporal evolution of the amplitudes $\lvert a_i(t) \rvert^2$ of the $2^7 = 128$ basis states $\Ket{\psi_i}$ this  system can be in. Again, all states are equally likely to be measured at $t=0$ but their amplitudes soon begin to diverge. In contrast to our first example, however, the amplitudes in this example do not evolve in a symmetric fashion. At $t=\tau$, there is indeed only a single basis state whose amplitude exceeds those of the other possible states. This state is $\Ket{0000001}$ and it yields the clustering result in Fig.~\ref{fig:example2}(b).

\section{Conclusion}

As of this writing, a growing number of reports predicts further rapid technological progress in the area of quantum computing \cite{Castelvecchi2017-QCR,Gibney2017-DWA,Economist2017-QD}. These anticipated developments will likely impact the field of machine learning, because quantum computers have the potential to accelerate the kind of optimization or search procedures that are at the heart of many machine learning algorithms.

In this paper, we were concerned with adiabatic quantum computing for machine learning. From an abstract point of view, the problem of setting up machine learning algorithms for adiabatic quantum computing can be understood as the problem of expressing their objective functions in terms of Ising models since existing adiabatic quantum computers are tailored towards solving these. Here, we applied this strategy to the problem of binary clustering and derived Ising models for $k=2$-means clustering. Two numerically simulated examples then demonstrated that the Schr\"odinger equations describing properly configured qubit systems can indeed be used for binary clustering of data.


\appendix

In this appendix, we prove our central claim in section~\ref{sec:k-means}, namely that, for a fixed sample $X$ of $n$ data vectors, the problem of solving
\begin{equation}
\amin{\vec{\mu}_1, \ldots, \vec{\mu}_k} S_W = \amin{\vec{\mu}_1, \ldots, \vec{\mu}_k} \sum_{i = 1}^k \, \sum_{\vec{x} \in X_i} \, \dsq{\vec{x}}{\vec{\mu}_i}
\end{equation}
is equivalent to solving
\begin{equation}
\amax{\vec{\mu}_1, \ldots, \vec{\mu}_k} S_B = \amax{\vec{\mu}_1, \ldots, \vec{\mu}_k} \sum_{i,j=1}^k n_i n_j \dsq{\vec{\mu}_i}{\vec{\mu}_j}.
\end{equation}

To this end, we consider the overall sample mean
\begin{equation}
\vec{\mu} = \frac{1}{n} \sum_{\vec{x} \in X} \vec{x}.
\end{equation}
and examine the total scatter of the data  
\begin{equation}
\label{eq:anova1}
S_T = \sum_{\vec{x} \in X} \dsq{\vec{x}}{\vec{\mu}}
\end{equation}
for which we note that it is a constant as long as $X$ is fixed. 

If the data in $X$ form $k$ clusters $X_i$ each with mean $\vec{\mu}_i$, we can rewrite this constant as
\begin{align}
S_T
& = \sum_{i=1}^k \sum_{\vec{x} \in X_i} \dsq{\vec{x}}{\vec{\mu}} \notag \\
& = \sum_{i=1}^k \sum_{\vec{x} \in X_i} \dsq{(\vec{x} - \vec{\mu}_i)}{(\vec{\mu} - \vec{\mu}_i)} \label{eq:anova2}
\end{align}
and the expression on the RHS of \eqref{eq:anova2} can be further expanded into a sum over three terms, namely
\begin{align}
T_1 & = \sum_{i=1}^k \sum_{\vec{x} \in X_i} \dsq{\vec{x}}{\vec{\mu}_i} \\
T_2 & = -2 \sum_{i=1}^k \sum_{\vec{x} \in X_i} \ipt{(\vec{x} - \vec{\mu}_i)}{(\vec{\mu} - \vec{\mu}_i)} \\
T_3 & = \sum_{i=1}^k \sum_{\vec{x} \in X_i} \dsq{\vec{\mu}}{\vec{\mu}_i} = \sum_{i=1}^k n_i \dsq{\vec{\mu}}{\vec{\mu}_i}.
\end{align}

The first of these terms is immediately recognizable as the conventional $k$-means clustering objective, that is $T_1 = S_W$.

For the second term, some straightforward algebra reveals
\begin{align}
T_2 
& = -2 \sum_{i=1}^k \Bigl( n_i \ipt{\vec{\mu}_i}{\vec{\mu}} - n_i \ipt{\vec{\mu}_i}{\vec{\mu}_i} - n_i \ipt{\vec{\mu}_i}{\vec{\mu}} + n_i \ipt{\vec{\mu}_i}{\vec{\mu}_i} \Bigr) \notag \\
& = 0
\end{align}
and for the third term we have
\begin{align}
T_3 
& = \sum_{i=1}^k \sum_{\vec{x} \in X_i} \Bigl( \ipt{\vec{\mu}}{\vec{\mu}} - 2 \, \ipt{\vec{\mu}}{\vec{\mu}_i} + \ipt{\vec{\mu}_i}{\vec{\mu}_i} \Bigr) \notag \\
& = n \, \ipt{\vec{\mu}}{\vec{\mu}} - 2 \, \vec{\mu}^T \sum_{i=1}^k n_i \, \vec{\mu}_i + \sum_{i=1}^k n_i \, \ipt{\vec{\mu}_i}{\vec{\mu}_i} \notag \\
& = n \, \ipt{\vec{\mu}}{\vec{\mu}} - 2 \, n \, \ipt{\vec{\mu}}{\vec{\mu}} + \sum_{i=1}^k n_i \, \ipt{\vec{\mu}_i}{\vec{\mu}_i} \notag \\
& = \sum_{i=1}^k n_i \ipt{\vec{\mu}_i}{\vec{\mu}_i} - n \, \ipt{\vec{\mu}}{\vec{\mu}}.
\end{align}

Next, we reconsider the between cluster scatter for which we find
\begin{align}
S_B 
& = \sum_{i,j=1}^k n_i n_j \dsq{\vec{\mu}_i}{\vec{\mu}_j} \notag \\
& = \sum_{i,j=1}^k \Bigl( n_i n_j \, \ipt{\vec{\mu}_i}{\vec{\mu}_i} - 2 \, n_i n_j \, \ipt{\vec{\mu}_i}{\vec{\mu}_j} + n_i n_j \, \ipt{\vec{\mu}_j}{\vec{\mu}_j} \Bigr) \notag \\
& = n \sum_{i=1}^k n_i \, \ipt{\vec{\mu}_i}{\vec{\mu}_i} - 2 \, n \, \ipt{\vec{\mu}}{\vec{\mu}} + n \sum_{i=1}^k n_j \, \ipt{\vec{\mu}_j}{\vec{\mu}_j}  \notag \\
& = 2 \, n \sum_{i=1}^k n_i \, \ipt{\vec{\mu}_i}{\vec{\mu}_i} - 2 \, n^2 \ipt{\vec{\mu}}{\vec{\mu}} \notag \\
& = 2 \, n \, T_3.
\end{align}

Putting all of this together, we therefore obtain
\begin{equation}
S_T = T_1 + T_3 = S_W + \tfrac{1}{2 \, n} S_B
\end{equation}
which, since $S_T$ and $n$ are constants, is to say that any decrease of $S_W$ implies an increase of $S_B$.

\bibliographystyle{IEEEtran}
\bibliography{literature}

\end{document}